%% file: nematzadeh.tex
\documentclass[alpha-refs]{wiley-article}

\usepackage{siunitx}

\papertype{Original Article}
\paperfield{Journal Section}

\usepackage{hyperref}
\renewcommand{\ref}[1]{\hyperref{#1}}
\usepackage{setspace}
\usepackage{url}
\usepackage{graphicx}
\usepackage{csquotes}
\usepackage{latexsym}
\usepackage{caption}
\usepackage{subcaption}
\usepackage{xcolor}

\newcommand\ie{\emph{i.e.}}
\newcommand\eg{\emph{e.g.}}

\newcommand{\ignore}[1]{}

\newcommand{\Table}[1]{Table~\ref{#1}}

\newcommand{\Example}[1]{Ex.~\ref{#1}}
\newcommand{\Figure}[1]{Figure~\ref{#1}}
\newcommand{\Equation}[1]{Eqn.~(\ref{#1})}
\newcommand{\Section}[1]{Section~\ref{#1}}
\newcommand{\Appendix}[1]{Appendix~\ref{#1}}

\newcommand\an[1]{\textcolor{blue}{#1}}

\newcommand{\rep}[3]{\mathrm{\theta}^{#1}_{#2#3}}

\DeclareMathOperator*{\argmax}{arg\,max}
\graphicspath{{plots/zero_unseen_prob/}}
\graphicspath{{plots/}}

\author[1\authfn{1}]{Aida Nematzadeh}
\author[2\authfn{2}]{Zahra Shekarchi}
\author[3\authfn{3}]{Thomas L. Griffiths}
\author[4\authfn{4}]{Suzanne Stevenson}

\contrib[1\authfn{1}, 2\authfn{2}]{Equally contributing authors.}
\blfootnote{1\authfn{1} Work was done prior to joining DeepMind.}

\affil[1]{DeepMind}
\affil[2]{University of Toronto}
\affil[3]{Princeton University}
\affil[4]{University of Toronto}

\corremail{nematzadeh@deepmind.com}

\runningauthor{Nematzadeh et al.}

\begin{document}
\title{Competition in Cross-situational Word Learning: A Computational Study}
\maketitle

\begin{abstract}
Children learn word meanings by tapping into the commonalities across different situations in which words are used and overcome the high level of uncertainty involved in early word learning experiences. We propose a modeling framework to investigate the role of mutual exclusivity bias -- asserting one-to-one mappings between words and their meanings -- in reducing uncertainty in word learning.
In a set of computational studies, we show that to successfully learn word meanings in the face of uncertainty, a learner needs to use two types of competition: words competing for association to a referent when learning from an observation and referents competing for a word when the word is used.
Our work highlights the importance of an algorithmic-level analysis to shed light on the utility of different mechanisms that can implement the same computational-level theory.
\end{abstract}
\clearpage

\clearpage

\input{01-intro.tex}
\input{020-related_work.tex}

\input{021-model-comp.tex}

\input{022-model-alg.tex}

\input{030-setup.tex}

\input{040-results.tex}

\input{050-conclusions.tex}

\newpage
\bibliography{nematzadeh}
\break
\input{060-appendix.tex}
\end{document}

%% file: 01-intro.tex
\section{Introduction}
\label{sec:intro}

One of the first steps in language acquisition is to learn word--meaning mappings, e.g., the word ``dog'' in the sentence ``see the dog'' refers to the tail-wagging animal under the kitchen table. This seemingly simple problem of word learning is a complex puzzle; in the initial phases of language development, children do not have any knowledge about word meanings (or other aspects of their language) and face a great deal of uncertainty. 
Without prior information, for a given word  (\eg, ``dog''), there is a high level of \textit{referential uncertainty} -- there are a great number of potential meanings in a child's environment (\eg, dog, table) that the word could refer to.
Similarly, there is high level of \textit{linguistic uncertainty} in mapping a referent (\eg, dog) to words in the utterances (\eg, ``see'', ``dog''). 
Moreover, an additional difficulty arises because not all the mappings between words and referents are one-to-one; sometimes words are mapped to more than one referent (\ie, homonyms such as ``bat'') or referents are mapped to more than one word (\ie, synonyms such as ``shut'' and ``close'').

Strong empirical evidence suggests that \textit{statistical cross-situational learning} helps both children and adults navigate these challenges, by gradually keeping track of statistical regularities across different situations (\eg, word--meaning co-occurrences), and using them to help resolve these ambiguous mappings \citep[\eg,][]{yu.smith.2007,smith.yu.2008}. 
However, cross-situational learning does not provide a detailed account of what mechanisms are responsible for resolving each type of uncertainty at different stages of word learning.

Moreover, a large body of developmental research has studied inductive biases that might facilitate word learning in the presence of different types of uncertainty \citep[\eg,][]{markman.1987}. A common theme among these biases is that competition can remove a number of possible hypotheses for a word meaning (and reduce the uncertainty). For example, the mutual exclusivity bias asserts that each referent is only mapped to one word \citep{markman.wachtel.1988}. 
This competition among referents means that given a new word and a number of possible referents, a learner reduces the uncertainty by not considering referents that are already associated to other words.   It is also suggested that such competitive processes play a role both locally and globally: there is competition when associating words and meanings from one observation (also known as \textit{in-the-moment learning}) as well as among all observed words and referents \citep[\eg,][]{yurovsky2013competitive}.\footnote{Work in computational modeling of cross-situational learning typically does not distinguish between the referent indicated by a word and its meaning.  We will use the terms \textit{referent} and \textit{meaning} interchangeably throughout this paper, while recognizing that there are important notions about the relations between those two that are being abstracted away by such an approach.}

Previous computational modeling work has shed light on the mechanisms and biases that might be involved in cross-situational learning \citep[\eg,][]{frank.etal.2007, fazly.etal.2010.cogsci,trueswell.etal.2013,nematzadeh.etal.2017.cogsci.bias}.
However, to our knowledge, no previous work has done an exhaustive analysis of the role of competition in the in-the-moment learning mechanisms and how these mechanisms interact with different representations of word meanings, 
which may also be influenced by competition.
In this work, our contributions are threefold: 
(1) We provide a general probabilistic formulation of cross-situational word-learning and show that the influential model of \citet{fazly.etal.2010.csj} is an instance of this formulation. 
(2) Using this formulation, we show how 
inductive biases can be modeled as competitive processes during in-the-moment and overall word learning, as well as comprehension of a word meaning. 
(3) Moreover, we examine how each modeling choice (and formulation of competition) affects learning in the presence of different sources of uncertainty, such as increased referential and linguistic uncertainty, fewer exposures to words, or acquiring homonyms and synonyms.

We find that the best model across all the tasks is the one that implements two types of competition, both among words and referents. Moreover, this competition happens during in-the-moment learning (for words) and comprehension (for referents). This result is different than previous modeling assumptions where a competition among referents was introduced during the overall learning of word meaning representations \citep{fazly.etal.2010.csj}. 
This finding suggests that the observed behavior in people \citep[\eg,][]{yurovsky2013competitive} might be explained by the competition during comprehension and not a global competitive process during learning.
We also observe that our best model performs better than the model of \citet{fazly.etal.2010.csj} in the presence of linguistic and referential uncertainty. Moreover, as opposed to the model of \citet{fazly.etal.2010.csj}, it can learn homonyms through implementing the competition mechanisms alone (without the need for any additional bias).

%% file: 020-related_work.tex
\section{Previous Work in Modeling Word Learning}
\label{sec:relatedwork}

\subsection{The Role of Computational Modeling}

Computational models have been used extensively to study various aspects of word learning in the last three decades \citep[\eg,][]{frank.etal.2007,alishahi.etal.2008.conll,kachergis.etal.2012.psyrev}. There are plenty of reasons to use computational models {in this area}: they provide a powerful tool to examine psycholinguistic theories of word learning, shed light on its underlying mechanisms, and investigate the interaction of different factors that might be involved.
We note that computational models of language acquisition are complementary to the experimental and theoretical studies: the predictions of already-verified models need to be examined in empirical studies. Moreover, these models can provide new directions for expanding the existing theories.

Several word learning models have been proposed, which address different aspects of the problem of learning word meanings. These models are built with specific assumptions, and use different input and learning algorithms.  
The models can be categorized into two broad groups based on their learning mechanism \citep{yu.smith.2012}:
(a) \textit{associative models} that argue word learning can be achieved through an associative learning mechanism -- by keeping track of associations between some entities/events. Connectionist models of word learning often belong to this category \citep[\eg,][]{regier.2005, colunga.smith.2005}. 
(b) \textit{hypothesis testing} models that are built based on the assumption that people keep track of multiple meaning hypotheses for each word, which are updated as people process new usages of that word. Most of these models are implemented using a Bayesian modeling framework \citep[\eg,][]{xu.tenenbaum.2007.psyrev}. However, some early rule-based approaches also belong to this group \citep[\eg,][]{siskind.1996}.  As is true of the learning mechanisms, the distinction between the two groups of models is not always clear, and their intersection is not necessarily empty. For example, the model of \citet{fazly.etal.2010.csj}
keeps track of hypotheses about word--referent pairs, similar to hypothesis
testing models, but also gathers co-occurrence statistics like associative
models. 

Another distinction between the models can be made based on the problem they investigate. A group of models focuses on single words, without considering the sentential context words occur in \citep[\eg,][]{regier.2005,colunga.smith.2005}. These models do not address the \emph{mapping problem}, \ie, learning to map each word in a multiword utterance to a referent that may or may not exist in the observed scene. On the other hand, given that the word--meaning mappings are known to these models, they examine other aspects of word learning such as predicting the inductive biases (such as shape bias) or patterns observed in child word learning (such as vocabulary spurt) \citep[\eg,][]{regier.2005,colunga.smith.2005,xu.tenenbaum.2007.psyrev,abbott.etal.2012}.
In the rest of this section, we summarize some of the models that address the mapping problem and learn the meaning of words from sentential context.

\subsection{Learning Words from Sentential Context}
\label{sec:comp-models-context}
The model of \citet{siskind.1996} is one of the first models that learns word meanings from an ambiguous context (including multiple words and multiple meanings). The model is rule-based and incremental: it learns mappings between words and their meanings by processing one input pair (an utterance of multiple words and its meaning representation) at a time, and applying a set of predefined rules to it. The predefined rules implement cross-situational learning and also encode word learning biases such as mutual exclusivity. \citet{siskind.1996} shows that his model can predict some of the observed patterns in child word learning such as learning homonymous words and fast mapping.
The main shortcoming of his approach is the inherent rigidity of the rule-based learning mechanism; follow-up models have turned to probabilistic learning mechanisms in order to better handle noise and uncertainty of the word learning input.

 The first probabilistic cross-situational word learning model is built by \citet{yu.ballard.2007}. Their model makes use of social cues such as the speaker's visual attention and prosodic cues in speech. The model
is an adaptation of the translation model of \citet{brown.etal.1993}: The speaker's utterances are considered as one language which is ``translated'' to a language consisting of the possible referents for words in the utterance. The input data consists of pairs of utterances and meaning representations, which are generated using two videos of mother-infant interactions taken from the CHILDES database \citep{macwhinney.2000}: the utterances are mother's speech represented as bags of words, and meaning representations are generated by
manually identifying objects presented in the scene when the corresponding utterance was heard. 
However, this model is not incremental: it performs a batch optimization process which iterates multiple times over the entire data, so it does not shed light on developmental patterns observed in child word learning.

To address this shortcoming, \citet{fazly.etal.2010.csj} developed an incremental model, which is also inspired by the translation model of \citet{brown.etal.1993}. However, as opposed to \cite{yu.ballard.2007}, who apply the translation model to their word learning data, \citet{fazly.etal.2010.csj} change the probabilistic formulation of the translation model.
Their model processes one input pair (an utterance represented as a bag of words and its scene representation consisting of a set of meaning symbols) at a time: For each word--meaning pair, it calculates an
\textit{alignment probability}  by probabilistically
aligning (mapping) the words (in the utterance) to the meaning symbols (in the
scene representation) using the current knowledge of word--meaning pairs. Then,
the stored knowledge of word--meaning pairs is updated using the new alignment
probabilities. For each word, the model learns a probability distribution, or
\textit{meaning probability}, over all possible meaning symbols, which
represents the model's current knowledge of that word. 
The utterances in the input are taken from the child-directed portion of the
CHILDES database. The scene representation for each utterance is generated
automatically, and is a set of meaning symbols corresponding to all words
in the utterance. These meaning symbols are taken from a gold-standard lexicon
in which each word is associated with its correct meaning. 
Although the scene representations are automatically generated, the input
resembles naturalistic child input in including linguistic and referential
uncertainty.
In addition, the input is reasonably large (around $170K$ input pairs), which makes it
possible to examine longitudinal learning patterns.  
\citeauthor{fazly.etal.2010.csj} perform a number of simulations, and show that the
model learns the meaning of the words under linguistic and referential uncertainty.
Furthermore, their model replicates several results of fast mapping experiments
with children, and can learn synonymous words. 

Several models have been proposed that improve the probabilistic cross-situational word learning framework introduced in \citet{fazly.etal.2010.csj} in various aspects. For example, \citet{kadar2015learning} use a more realistic input data where referents are represented by images; similarly, \citet{lazaridou2016multimodal} also propose a multi-modal account of semantic learning.
\citet{nematzadeh.etal.2013} model the acquisition of a single meaning for non-literal multi-word expressions (\eg, ```give a knock on the door''). Other work integrates attention and forgetting mechanisms into the word learning process \citep[\eg,][]{nematzadeh.etal.2012.cmcl,kachergis.etal.2012.psyrev}. \citet{vong2020learning}, using a self-supervised approach, introduce an account of cross-situational word learning from images and words (at the computational level). In the rest of this section, we explain two other families of word learning models.

\citet{stevens.etal} propose an incremental model that builds upon the word learning experiments of \citet{medina.etal.2011} and \citet{trueswell.etal.2013}; the authors argue that people only maintain a single meaning hypothesis for each word. They claim that keeping track of cross-situational statistics for word--meaning pairs is not necessary in word learning.
Their computational model implements a probabilistic version of the single
meaning hypothesis.  It calculates an association score for each
word--meaning pair which is very similar to the score calculated in
\citet{fazly.etal.2010.csj}.  The key difference is that for a given word,
only the association score of the most likely meaning is updated (as opposed to
the model of \citet{fazly.etal.2010.csj} that updates the score of all the
meanings observed with a word).
The authors compare the performance of their model with a few other models
(including the models of \citet{frank.etal.2009} and
\citet{fazly.etal.2010.csj}). They train each model on a dataset of
child-directed utterances paired with manually-annotated scene representations.
They show that their model outperforms other models in learning a lexicon.
However, their dataset is very small (less than 1000 utterances). Moreover,
it does not include much referential uncertainty because only a subset of words
in utterances (concrete nouns) are annotated in the scene representations and
considered in the evaluations.
Thus, it is not clear whether their model would perform as well on a larger and
more naturalistic dataset.

\citet{frank.etal.2009} propose a Bayesian model of word learning and speakers' communicative intentions:
the speaker intends to talk about a subset of objects he observes, and uses some words to express this set of objects.  Given a corpus of situations consisting of such words and objects, the goal of the model is to find the most probable lexicon. The model favors a smaller lexicon, which is encoded in the formulation of the prior probability. This choice of prior enforces a conservative learning approach, in which learning all the existing word--object pairs is not a priority. In calculating the likelihood, all intentions (subsets of objects) are considered equally likely. Thus, the model is not incorporating a fully elaborated model of speaker's communicative intentions. The lexicon with the maximum a posteriori probability is chosen by applying a stochastic search on the space of possible lexicons. The model of \citet{frank.etal.2009} replicates several patterns observed in child word learning, such as the mutual
exclusivity bias and fast mapping. The input data is generated using the same videos of mother--infant play time that \citet{yu.ballard.2007} used. The meaning representations are similarly produced, by manually hand-coding all the objects that were visible to the infant upon hearing each utterance. Although the data is very similar to children's possible input, the size of the data set is very small; moreover, the learning in the models is not incremental, which makes certain longitudinal patterns (\eg, vocabulary spurt) impossible to examine.

\subsection{Summary} 

Several models address the mapping problem and learn words from context \citep{siskind.1996, yu.ballard.2007,frank.etal.2009,fazly.etal.2010.csj, kachergis.etal.2012.psyrev}. Among these, \citet{yu.ballard.2007} and \citet{frank.etal.2007} use naturalistic child data, in which utterances are taken from caregivers' speech and the meaning representations for each utterance are generated by manually annotating the objects and social cues in the environment. The shortcoming of this
approach in data generation is that the resulting dataset is very small. To address this, \citet{fazly.etal.2010.csj} take the utterances from caregivers' speech but automatically generate the meaning representations. By doing so, they have the advantage of simulating experiments or observational studies that examine longitudinal patterns of word learning. The presented models also differ in their incorporation of word learning biases and constraints. Moreover, the learning algorithms of some of the models are incremental; thus, they are more similar to child word learning \citep[\eg,][]{siskind.1996,fazly.etal.2010.csj}. In contrast, the learning algorithms of most of the models are batch processes, and process all the input at once \citep[\eg,][]{xu.tenenbaum.2007.psyrev,
frank.etal.2009}.

%% file: 021-model-comp.tex
\section{A Computational-level Model of Word Learning}
\label{sec:complevel}

Research at the computational level of analysis \citep{marr.1982} focuses on specifying the goal of the problem of word learning. There are two main approaches in defining the problem:  (1) finding a lexicon that maximizes the likelihood of a corpus \citep[\eg,][]{frank.etal.2007}, or (2) treating word learning as a translation problem \citep[\eg,][]{yu.ballard.2007} -- while in a typical machine translation setting, the goal is to translate the sentences from a source to a target language (\eg, French to English), in word learning a model learns to translate between words and a set of referents (possible meanings).\footnote{Here, we use the terms referent and meaning interchangeably.} 
\citet{fazly.etal.2010.csj} propose an incremental model that implements the same translation idea but makes it possible to study different patterns observed at various stages of development. This work has inspired many extensions and follow-up models \citep[\eg,][]{nematzadeh.etal.2012.cmcl,cassani2016constraining} because of its incrementality and also as it provides a formulation for both in-the-moment and overall learning.
However, this model has a fixed set of assumptions about the implemented in-the-moment learning mechanism and its word-meaning representation. Although these assumptions (adopted from the machine translation literature) result in a robust model that predicts a wide range of behaviors, they are not grounded in the word learning literature or compared to alternatives.
We first propose a general probabilistic formulation for cross-situational learning and show how the model of \citet{fazly.etal.2010.csj} (henceforth, FAS) can be derived from that formulation.
This formulation can be considered as a computational level solution to the cross-situational learning problem.
We then show how different types of competition mechanisms can be introduced into the probabilistic formulation of the model. Similar analysis can be applied to other computational-level theories; we choose the cross-situational learning formulation since it has successfully predicted many word learning behaviors observed in children and has minimal assumptions about the problem.

\subsection{Word Learning Input}
The first step in modeling word learning is to simulate a child's input, \ie, what she hears and perceives in her environment. We assume each utterance {(what is heard)} is a set of words, and do not include information about word order. Following previous work \citep[\eg,][]{fazly.etal.2010.csj}, we use a set of referents to represent the word learning scene -- what a child perceives while hearing an utterance. Our word learning input is then a sequence of such utterance--scene pairs as shown in \Example{US-features}.  
\vspace{-.3cm}
\begin{eqnarray}
{\small
\begin{tabular}{l}
{\bf Utterance:} $\{$\emph{Ray}, \emph{eats}, \emph{an}, \emph{apple}$\}$ \\
{\bf Scene:} $\{$ \textsc{ray}, \textsc{eats}, \textsc{an}, \textsc{apple}$\}$ \\
\end{tabular}}
\label{US-features}
\vspace{-.3cm}
\end{eqnarray}
\noindent
Note that the input does not encode what referents belong to the meaning of each word; as a result, the input is similar to the natural data perceived by children with respect to \textit{alignment ambiguity} -- \ie, the learning algorithm must solve the word--referent mapping problem.

\subsection{Probabilistic Formulation}

We assume our observations create a corpus $C$ of utterance-scene ($u$--$s$) pairs and $C$ is generated by sampling from an unknown distribution $\theta$. $\theta$ relates words and referents, and represents our probabilistic knowledge of word meanings. 
The word learning problem is then to estimate $\theta$ given the corpus $C$. $\theta$ is treated as a parameter, and we find the $\theta$ that best explains our observations, $C$. 
In this regard, we assume that a hidden variable, $a$, called the alignment, determines how words and referents are mapped (aligned) in a given $u$--$s$ pair. Given $a \in A$, where $A$ is a set of all possible alignments, the likelihood of our observations can be defined as:
\begin{align}
\vspace{-.2cm}
p(C| \theta)  &= \prod_{\substack{u,s \in C}} p(u,s| \theta) \nonumber \\
              &= \prod_{\substack{u,s \in C}} \sum_{a \in A} p(a, u, s| \theta) 
\vspace{-.3cm}
\label{eq:cll}
\end{align}
In \Equation{eq:cll}, the first line defines the likelihood of our observations $C$ (given $\theta$) in terms of individual utterance-scene pairs. We assume that $u$-$s$ pairs are independent. Therefore, the likelihood is the product of the probability of $u$-$s$ pairs.\footnote{In reality, utterances in a conversation are often dependent as they follow a specific topic. Here, we assume that young children attend more on each utterance and may ignore the sequentiality of utterances. This simplifying assumption also makes the calculation of the probabilities easier.} The second line specifies how the probability of each $u$-$s$ pair can be calculated through the hidden variable $a$, which determines how words and referents, in the $u$-$s$ pair, are mapped. The assumptions about the structure of alignment $a$ (\eg, allowing only one-to-one mappings) and calculation of $p(a,u,s|\theta)$ result in models with different inductive biases.

To estimate $\theta$, we use the incremental version of the Expectation Maximization (EM) algorithm \citep{neal.hinton.1998}. The incrementality of this algorithm enables the model to update the word meanings ($\theta$) after processing each $u$-$s$ pair. 
In the absence of any prior knowledge, all referents are equally likely to correspond to a word. Therefore, to start with an initial value for $\theta$, we assumes a uniform distribution over referents and then iteratively repeat the following steps. 
\begin{enumerate}
    \item \textbf{Calculate the alignment probability.} For each utterance, we calculate the probability of each possible alignment given the current $\theta$ -- how strongly each word--referent pair is currently associated. This corresponds to the Expectation step of the EM algorithm. The probability of the alignments for each $u$-$s$ pair is estimated:
    \begin{align}
    \vspace{-.3cm}
    p(a|u, s,\theta) &= \frac{p(a, u, s| \theta )}{p(u, s| \theta)}  = 
    \frac{p(a, u, s| \theta)}{\sum\limits_{a \in A}p(a, u, s| \theta)} 
    \label{eq:emalign}
    \vspace{-.3cm}
    \end{align}
    In the following sections, we will explain how the right-hand side of \Equation{eq:emalign} is calculated differently for each model depending on its assumptions about the alignment variable. (For example, \Equation{eq:emfasalign} is one instance where referents in the scene are assumed to be independent.)

    \item \textbf{Update $\theta$.} Given the newly calculated alignment probabilities and the value of the parameter at the previous time step, $\theta^{t-1}$, we update $\theta$ such that it maximizes the likelihood of the observations, $C$. At the Maximization step, given the value of the parameter at the previous time step, $\theta^{t-1}$, the current $\theta$ is selected to maximize the following:
    \begin{align}
    \vspace{-.2cm}
    \hat{\theta} &= \argmax_{\theta} \textrm{E}_{A|C, \theta^{t-1}}[ p(C, A|\theta)]
    \label{eq:emmaxstep}
    \vspace{-.2cm}
    \end{align}
    where $\textrm{E}_{A|C, \theta^{t-1}}$ is the conditional expectation of $p(C, A|\theta)$ -- i.e., the expected value of corpus $C$  and the alignment variables $A$ given their probability $\theta$ (which was calculated at the previous time step). 
\end{enumerate}

Given this general formulation, we can derive different models by enforcing different constraints on $\theta$ and $a$; specifically, either can be a joint or conditional probability distribution.  The former adds dependence and thus competition between all words and referents. The latter, depending on the direction of conditional probability, either introduces competition among words for a referent, $p(w|r)$, or among referents for a given word, $p(r|w)$.  
Note that the different assumptions about the alignment variable $a$ produces different in-the-moment learning mechanisms, with different {competition} processes. In addition, different formulations of $\theta$ produce different overall learning mechanisms. We next derive the FAS model using our probabilistic formulation and explain the assumptions behind this model.
\subsection{The FAS Model}
FAS propose a formulation to calculate the alignment probability (\Equation{eq:emalign}) and $\theta$; however, they do not discuss what the probabilistic model behind these formulas is and how they can be derived. We define the FAS model in terms of the probabilistic framework introduced in the previous section. Here we provide a high-level outline of how the FAS model can be derived, and explain the detailed derivation in \Appendix{sec:appendix}.

Recall that each scene representation $s$ constitutes of a set of referents $r_i$. The FAS model assumes that given an utterance $u$, the referents are generated independently; so, instead of calculating $p(s,u|\theta)$ as in \Equation{eq:cll}, the likelihood is defined over the conditional probability of referents given an utterance:
\begin{align}
\vspace{-.2cm}
p(C| \theta)  &= \prod_{\substack{u,s \in C\\r_i \in s}} p(r_i|u, \theta) \nonumber \\
&= \prod_{\substack{u,s \in C\\r_i \in s}} \sum_{a \in A} p(r_i, a|u, \theta) 
\vspace{-.2cm}
\end{align}
Here, the alignment variable $a$ defines the mappings between words and a given referent. Given the utterance $u$, it determines both the word ($w_j$) and its strength of mapping to a given referent $r_i$. We use the notation $a_j$ to refer to the alignment determines the mapping to $w_j$. Thus, the value of $a_j$ determines the strength of the mapping between $w_j$ and $r_i$.
\begin{align}
\vspace{-.2cm}
p(a_{j}|u,r_i,\theta) &= \frac{p(a_{j},r_i| u, \theta)}{p(r_i|u, \theta)} = \frac{\theta_{ij}} {\sum_{w_k \in u} \theta_{ik}} 
\label{eq:emfasalign}
\vspace{-.2cm}
\end{align}
where $\theta_{ij}$ specifies the association of $w_j$ with the referent $r_i$, given the learned distribution $\theta$. Note that this corresponds to the Expectation step of the EM algorithm, and \Equation{eq:emfasalign} is an instantiation of \Equation{eq:emalign}.

In the Maximization step, the new value of $\theta$ is calculated by finding $\theta$ that maximizes the model's likelihood as given in \Equation{eq:emmaxstep}. The FAS model assumes that $\theta$ is a conditional probability, $p(r|w)$. This suggests that there is an additional dependence assumption about the learned representations: each word is a distribution over the referents, and thus given a word, the referents compete to be associated with that word.
To find $\theta$, the derivative of the likelihood is calculated and equated to zero. This results in:\footnote{The details of this derivation are given in \textbf{\Appendix{sec:appendix}}.}
\begin{align}
\vspace{-.2cm}
p(r_i|w_j)= \frac { p(a_{j}|u,r_i, \theta^{t-1})\ \mathrm{count}(r_i, w_j)} 
{\sum\limits_{r_m \in S}  p({a_j}|u,r_m, \theta^{t-1})\ \mathrm{count}(r_m, w_j)}
\vspace{-.2cm}
\end{align}
where $w_j$, determined by the alignment variable $a$, is the word mapped to $r_i$. also, $count(r_i,w_j)$ is the number of times $r_i$ and $w_j$ have co-occurred in the corpus $C$. We can approximate $p(a_{j}|u,r_i, \theta^{t-1})\ \mathrm{count}(r_i, w_j)$ by adding the current alignment probability, $p(a_{j}|u,r_i, \theta^{t-1})$, to the sum of all the previously calculated 
ones -- instead of multiplying it to the number of times the word and referent co-occur. Note that the value of the current and previous calculated alignment probabilities can be different.
This approximation enables us to calculate $p(a_{j}|u,r_i, \theta^{t-1})\ \mathrm{count}(r_i, w_j)$  incrementally. In order to do this, FAS defined an association score, $assoc$. $assoc$ is updated as the model processes $u$-$s$ pairs at each time step:
\vspace{-.1cm}
\begin{eqnarray}
\mathrm{assoc}_{t}(w_j,\,r_i) = \mathrm{assoc}_{t-1}(w_j,\,r_i) +  a_t(w_j|r_i)
\label{eq:assoc}
\vspace{-.2cm}
\end{eqnarray}
where $a_t(w_j|r_i)$ equals $p({a_{j}}|u,r_i, \theta^{t-1})$, and the initial value of $\mathrm{assoc}(w_j,\,r_i)$, before the first co-occurrence of $w_j$ and $r_i$, is zero.  
Intuitively, $assoc$ score represents the overall association strength of a word and a referent, and it captures how strongly the word-referent pair is associated in each observation, $u$-$s$.

%% file: 022-model-alg.tex
\section{Algorithmic-level Analysis of Word Learning}

An algorithmic-level analysis \citep{marr.1982} focuses on the effect of choices of the algorithms and representations. \citet{Dupoux_2018} and \citet{tsutsui2020computational} also argue that to understand how infants learn language in the real world, we need to model internal learning processes. Here, we perform an algorithmic-level investigation of cross-situational word learning. We study the following questions: What are the possible in-the-moment learning mechanisms? Given exposure to language over time, how are the word meanings represented in semantic memory? Or, how is in-the-moment evidence accumulated to formulate a word meaning? How is a word meaning used in different tasks?  And what is the role of competition in each of these steps of learning?

We study these questions by using our general cross-situational word learning formulation. A learner starts with some assumptions about the word-meaning representation -- the distribution, $\theta$, relates words and referents. The initial value of $\theta$ is equal and uniform across all word--referent pairs. 
The learner uses its current $\theta$ and in-the-moment learning algorithm to update the probability of alignments -- how words and referents are mapped given the current input pair.
The model updates word meanings, $\theta$, using the new alignments.

\subsection{In-the-Moment Learning Mechanisms}

We formulate different in-the-moment learning mechanisms by varying the definition of the alignment variable (\Equation{eq:emalign}), which associate words and referents in each given input. One possibility is that the alignments do not interact -- each alignment of a word and a referent is calculated independently of all others; in other words, the increase or decrease in the strength of one alignment does not affect any other. This definition assumes that a learner only uses its previous knowledge of the word and referent alignment. So, other available information, such as what other words and referents are in the input, does not bias its assessment. As a result, we can formulate the alignment variable as follows:

\vspace{-1.5cm}
\begin{eqnarray}
a_{t}(w,r) ={\rep{t}{w}{r}}
\hspace{.5in} \mathrm{[No\ Bias]}
\label{eq:alignwandf}
\end{eqnarray}

However, it should be mentioned that experiments in developmental psychology show that children make use of this information in determining meanings of words. For example, when young children hear a novel word (\eg, ``dax''), while presented with a known object (for which they have heard a label) and an unfamiliar object, they intend to choose the unknown object as the referent of the novel word \citep[\eg,][]{markman.wachtel.1988}. This tendency is often referred to as the \textit{mutual exclusivity bias} -- \ie, to limit the number of words for a meaning in this example. Similarly, we can consider the other direction of this bias, where the number of meanings associated to a word is limited; such an effect of dis-preferring homonymy has also been observed in children \citep[\eg,][]{casenhiser2005children}.

 The independence assumption of the alignment variables can be discarded, such that there is competition (dependence) among words for a given referent, or among referents for a given word. We implement the first competition of words for a referent (word competition; WC) by defining a distribution over words for each referent (\ie, by normalizing \Equation{eq:alignwandf} over the words in the utterance):
\vspace{-.1cm}
\begin{eqnarray}
\mathit{a_{t}(w|r) = \displaystyle\frac{\rep{t}{w}{r}}{\displaystyle\sum_{\mathbf{w' \in \mathrm{U}_t}}\rep{t}{w'}{r}}}
\hspace{.5in} \mathrm{[WC]}
\label{eq:alignwgivenf}
\end{eqnarray}
\noindent Note that this is the formulation of alignment probability that the FAS model employ, \Equation{eq:emfasalign}.

We can similarly model the other direction of competition by introducing dependence among referents given a word (referent competition; RC):
\vspace{-.1cm}
\begin{eqnarray}
\mathit{a_{t}(r|w) = \displaystyle\frac{\rep{t}{w}{r}}{\displaystyle\sum_{\mathbf{r' \in \mathrm{S}_t}}\rep{t}{w}{r'}}}
\hspace{.5in} \mathrm{[RC]}
\label{eq:alignfgivenw}
\end{eqnarray}

\subsection{Updating the Word Meanings} 

The second stage of learning involves defining and updating the word meaning representations. Following \citet{fazly.etal.2010.csj}, we calculate an association score, $\mathrm{assoc}_{t}(w,\,r)$ -- to capture the strength of association of a word and a referent. It should be pointed out that $\mathrm{assoc}_{t}(w,\,r)$ is the sum of their alignments that is incrementally accumulated over time, as in \Equation{eq:assoc}.\footnote{This formulation assumes the learner has a perfect memory, but this assumption can be relaxed by letting alignments decay over time \citep[\eg,][]{kachergis.etal.2012.psyrev,nematzadeh.etal.2012.cmcl}.}
Depending on the given in-the-moment learning mechanism, the alignment probability is calculated using one of the \Equation{eq:alignwandf}, \Equation{eq:alignwgivenf}, or \Equation{eq:alignfgivenw}. 

One possibility to update $\theta$ is to assume word--referent associations are independent in their contribution to the learned representation of a word. Based on this assumption, a ``No Bias'' word meaning can be represented as:\footnote{Although the $\mathrm{assoc}$ score is not normalized and thus is not a probability, for simplicity, we refer to it as $p(w,r)$. We could also consider $p(w,r)$ to be a joint distribution, and normalize $\mathrm{assoc(w,r)}$ over both referents and words. However, this normalization would cancel out in the numerator and denominator of two of the alignment formulations, \Equation{eq:alignwgivenf} and \Equation{eq:alignfgivenw}. Interestingly, when there is a competition during in-the-moment learning, the joint probability and unnormalized association scores (\Equation{eq:probwandf}) are the same models. They are only different if we use $a(w,r)$, \Equation{eq:alignwandf}.}
\begin{eqnarray}
\rep{t+1}{w}{r} = \mathit{p_{t+1}(w,r) = \mathrm{assoc}_{t}(w,\,r) }
\hspace{.20in} \mathrm{[No\ Bias]}
\label{eq:probwandf}
\vspace{-.1cm}
\end{eqnarray}

Nonetheless, we can consider dependence among referents (referent competition -- RC), where a word meaning representation is a distribution over all observed referents, $p(\cdot|w)$. To put it differently, we consider that \textit{all} the observed referents compete to be associated with a word. Thus, this competition sharpens the connection of a word with its referent in the meaning representation:
\vspace{-.1cm}
\begin{eqnarray}
\rep{t+1}{w}{r} = \mathit{p_{t+1}(r|w) = \displaystyle\frac{\mathrm{assoc}_{t}(w,\,r)}
          {\displaystyle\sum_{r'\in \mathcal{M}}{\mathrm{assoc}_{t}(w,\,r')}}} 
\hspace{.20in} \mathrm{[RC]}
\label{eq:probfgivenw}
\vspace{-.1cm}
\end{eqnarray}
where $\mathcal{M}$ is the set of all referents observed up to time $t$. Note that this formulation introduces a global competition (as opposed to a local competition that is introduced during in-the-moment learning). The global competition results in a stronger bias than the one implemented during the in-the-moment learning, because all the observed referents compete for a given word (as opposed to the competition among referents that occur in a given scene). This bias helps the model associate a word to \textit{relevant} referents (that consistently co-occur with it) better than the \textit{irrelevant} ones. 
This \Equation{eq:probfgivenw} and \Equation{eq:alignwgivenf} make the FAS model.

Logically in $\rep{t}{w}{r}$, we could also model the competition among words for a given referent, analogous to this direction of competition in the $a_t$ function.  However, the resulting representation, $p(w|r)$, does not yield a straightforward meaning representation for each word. It instead provides a representation for each referent (i.e., a distribution over the words that refer to it). In fact, when we evaluate the models on how well they estimate $p(w|r)$, not surprisingly their performance is the same as their mirror models; $a(r|w)p(r|w)$ performs the same as $a(w|r)p(w|r)$.\footnote{The models are like each other if we assume words and referents are swapped. Words are referents and referents are words. Because the utterances and the scenes are the same without referential uncertainty, these models are equal.}

We do not report these results in this paper, and leave it for the future to investigate for which tasks a referent representation is a better choice (as opposed to a word representation). 

Taking everything into account, we examine six models; by choosing from all combinations of the aforementioned in-the-moment learning mechanisms and word-meaning representations; see \Table{tb:models}. We discuss how comprehension of word meanings can introduce competition in the next section.

\begin{table}
\centering
{
\begin{tabular}{ll|cc|cc}
 &  &  \multicolumn{2}{c|}{In-the-moment learning} & \multicolumn{2}{c}{Meaning rep.}\\
 & Model   & Bias   & Eqn.\#    & Bias   & Eqn.\#\\
\hline
1. & $a(w,r)p(w,r)$   &  No Bias & \Equation{eq:alignwandf}   & No Bias  & \Equation{eq:probwandf}\\
2. & $a(w|r)p(w,r)$   &  WC & \Equation{eq:alignwgivenf}   & No Bias  & \Equation{eq:probwandf}\\
3. & $a(r|w)p(w,r)$   &  RC  &  \Equation{eq:alignfgivenw}   & No Bias  & \Equation{eq:probwandf}\\
4. & $a(w,r)p(r|w)$   &  No Bias  &  \Equation{eq:alignwandf}   & RC  & \Equation{eq:probfgivenw}\\
5. & $\mathbf{a(w|r)p(r|w)}$   &  \textbf{WC}  & \textbf{
 \Equation{eq:alignwgivenf}}   & \textbf{RC}  & \textbf{\Equation{eq:probfgivenw}}\\
6. & $a(r|w)p(r|w)$   &  RC & \Equation{eq:alignfgivenw}   & RC  & \Equation{eq:probfgivenw}\\
\end{tabular}
}
\caption{List of all the models. The model marked in bold is the same as the FAS model. RC stands for referent competition and WC stands for word competition.}
\label{tb:models}
\end{table}

\subsection{Comprehending the Word Meanings}

In previous sections, we have discussed how a learner can maintain a representation for each word meaning, and update this representation after processing each new observation (including the word). In this section, we explain how the comprehension of the word meanings can be formulated.

In typical comprehension experiments \citep[\eg,][]{yu.smith.2007}, a word is heard while some referents -- including the intended referent for that word -- are present. It is assumed that the learner comprehend the word if they pick its correct referent. In this set-up, it is clear that during comprehension, referents compete for a given word -- the learner will pick the referent whose representation is the most similar to the learner's representation of the word.

The same competition also happens in everyday instances of comprehension; in other words, given a word and a possible set of referents, the learner picks a referent whose meaning best resembles the learned meaning of the word. However, in this case it is not clear what constitutes the possible set of referents; \eg, is it all the referents the learner knows? Here, without defining a set of competing referents, we introduce this referent competition in the measure of a learner's overall comprehension of words.

Intuitively, a learner can understand a given word if their representation of the word (\ie, the learned word meaning) is similar to the actual representation of the referent which was intended by the speaker uttering the word. Let's consider a situation in which a speaker says the word ``dax''; the listener comprehends the meaning of ``dax'' if the intended meaning of ``dax'' is similar to the listener's learned representation of ``dax''. It is assumed here that the intended meaning  corresponds to some perceived referent in the world, which we represent by the gold-standard meaning of the word.

We thus define the comprehension of a word as the similarity of the learned and gold-standard representations. Based on this definition, we use a very simple yet common similarity measure -- the cosine similarity -- to introduce few assumptions on the formulation of the comprehension score.
More specifically, following \citet{nematzadeh.etal.2012.cogsci}, at any timestep $t$, we calculate a \textit{comprehension score} for each word $w$ by comparing its learned meaning representation, $\text{lrnd}(w)$, with the gold-standard meaning of the word, $\text{gold}(w)$, using cosine similarity:
%
\begin{equation}
    \mathrm{comp}_t(w) = \text{cosine}(\text{lrnd}(w), \text{gold}(w))= 
    \frac{\text{lrnd}(w) \cdot  \text{gold}(w)}{\|\text{lrnd}(w) \|\|\text{gold}(w) \|}
    \label{eq:acq-score}
\end{equation}
\noindent 
where $\text{gold}(w)$ is a vector containing all the semantic referents. So, the value of each element in the vector is 1 if a referent is the gold-standard meaning of $w$ and 0 for other irrelevant referents. $\text{lrnd}(w)$ is a similar vector in which the strength of $w$ and a referent is taken from $\theta$. The comprehension scores can be averaged over all observed words to obtain the \textit{average comprehension score}. This average comprehension score shows the overall knowledge of a learner at time $t$.

Note that using cosine as a similarity measure introduces a competition among referents. Because, the denominator is a multiplication of norms of two vectors $\text{lrnd}(w)$ and $\text{gold}(w)$ which introduces dependence and therefore competition among all referents for $w$.\footnote{Our primary experiments with other similarity measures show that using cosine improves the performance across all models. We leave an extensive analysis of the effect of these measures to future work.}

%% file: 030-setup.tex
\section{Simulations}
In this section, we first discuss how the input data for our simulations is generated. Then we study the role of two types of competition (among referents and among words) at different stages of learning and comprehension:
\begin{itemize}
    \item \emph{In-the-moment learning}, which introduces a local competition among words and referents in a given observation. 
    \item \emph{Overall learning}, which is equivalent to a global competition among words and referents in the word meaning representation. 
    \item \emph{Comprehension of word meaning}, which shows how similar learned and gold-standard representations of a word are.
\end{itemize}
Moreover, we examine how each of these conditions performs in word learning situations that differ in terms of their difficulty; what type of competition (if any), and at which stage, makes the learner more robust in the presence of uncertainty?

\subsection{Simulation Input Data}
\label{sec:setup}

The utterances in the input to the model are taken from child-directed speech in the Manchester corpus \citep{theakston.etal.2001} of CHILDES \citep{macwhinney.2000}\footnote{The Manchester corpus contains transcripts of conversations of caretakers with 12 children between the ages of 20 and 36 months.}. Since there is no data about semantic representations of the scenes in which these conversations occurred, similar to \citet{fazly.etal.2010.csj}, we generate the corresponding meaning symbols to the words in an utterance as a set of correct referents. Therefore, each utterance in the corpus is input to the model as an unordered set of words, and each scene is represented as a set of referents corresponding to those words in the utterance (see \Example{US-features})\an{\footnote{Here we consider one-to-one mappings between words and their referents.  We use a different setup (in \Section{sec:LearningSynonymsAndHomonyms}) when we study tasks related to homonyms or synonyms.}}. 
The input corpus $C$ is a sequence of $N$ 
utterance--scene pairs $u$--$s$.\footnote{In our simulations here, we set $N=6,000$ because all the models reach a stable performance well before then.} Each input pair $u_i$--$s_i$ in $C$ basically has minimal referential uncertainty (RU) or linguistic uncertainty (LU):  Uncertainty is imposed by the ambiguity of the alignment problem (that which word maps to which meaning). But, there are no ``extra'' meanings in $s_i$ which are not associated with a word in $u_i$ (increasing the RU), nor are there ``extra'' words in $u_i$ which have no corresponding referents in $s_i$ (increasing the LU).  For creating our various input corpora, we start with this sequence $C$ and further process it to create $3$ types of corpora with varying amounts of RU and LU to more accurately simulate the difficulty of ambiguous learning environments.

To create an input corpus with increased referential uncertainty, we use every third input pair from $C$ as the basis for an input pair, and use the in-between input pairs for adding ``extra'' referents to the scene.  We call this input corpus RU+; see column RU+ in \Table{table:RULUcorpus} for the first three input pairs. 
Analogously, to create data with increased linguistic uncertainty, we again use every third input pair from $C$, and use the in-between input pairs for adding ``extra'' words to the utterance.  See column LU+ in \Table{table:RULUcorpus} for the first three input pairs.
Now to form an input corpus having minimal RU and LU that is consistent with the same core utterance and scene pairs as in the RU+ and LU+ corpora, we use every third input pair from the original sequence $C$, as shown in Input column in \Table{table:RULUcorpus}. By using in-between pairs to get referents (words) to add referential (linguistic) uncertainty, we maintain the context of the utterances, as apposed to adding  random referents (words) as in \citet[][]{siskind.1996}.
\begin{table}[h!]
\centering
\small{
\begin{tabular}{||c l l l ||} 
\hline
 Time & Input & RU+ & LU+ \\ [0.5ex] 
\hline\hline
1 & [$u_1$, $s_1$] & [$u_1$, $s_1$+$s_2$+$s_3$] & [$u_1$+$u_2$+$u_3$, $s_1$] \\ 
\hline
2 & [$u_4$, $s_4$] & [$u_4$, $s_4$+$s_5$+$s_6$] & [$u_4$+$u_5$+$u_6$, $s_4$] \\
\hline
3 & [$u_7$, $s_7$] & [$u_7$, $s_7$+$s_8$+$s_9$] & [$u_7$+$u_8$+$u_9$, $s_7$] \\
\hline
\end{tabular}
}
\caption{Three types of input corpora: with no added RU/LU (Input), with increased referential uncertainty (RU+), and with increased linguistic uncertainty (LU+).}
\label{table:RULUcorpus}
\end{table}

%% file: 040-results.tex
\subsection{Simulation Results}
\label{sec:results}

We first analyze the overall learning performance of the six models introduced in \Table{tb:models}. Then we study how each model performs in more challenging situations similar to what children face when learning word meanings. We examine the role of different sources of uncertainty in forming the word--referent mappings: referential uncertainty, linguistic uncertainty, limited exposure to words, and learning homonyms or synonyms.

\subsubsection{Overall Learning Patterns}

\label{sec:overallpatterns}
\begin{figure}
\centering{
\begin{subfigure}{0.56\textwidth}
\includegraphics[width=\linewidth]{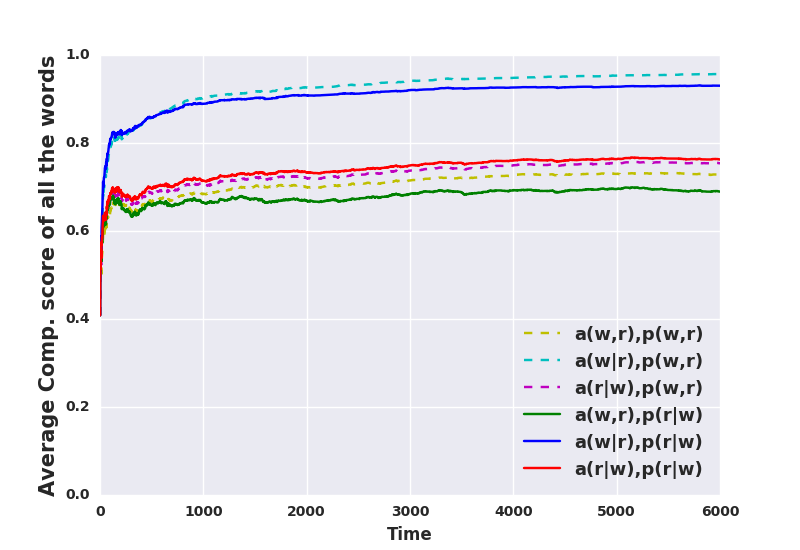}
\caption{Default input corpus: no added RU or LU.}
\label{fig:avgcomp-noru-nolu}
\end{subfigure}
\begin{subfigure}{0.56\textwidth}
\includegraphics[width=\linewidth]{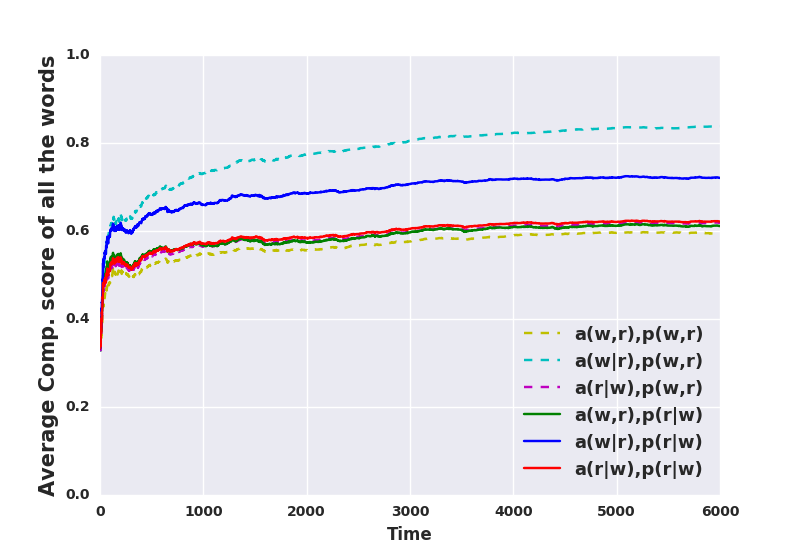}
\caption{RU+ input corpus.}
\label{fig:avgcomp-ru2-nolu}
\end{subfigure}
\begin{subfigure}{0.56\textwidth}
\includegraphics[width = \linewidth]{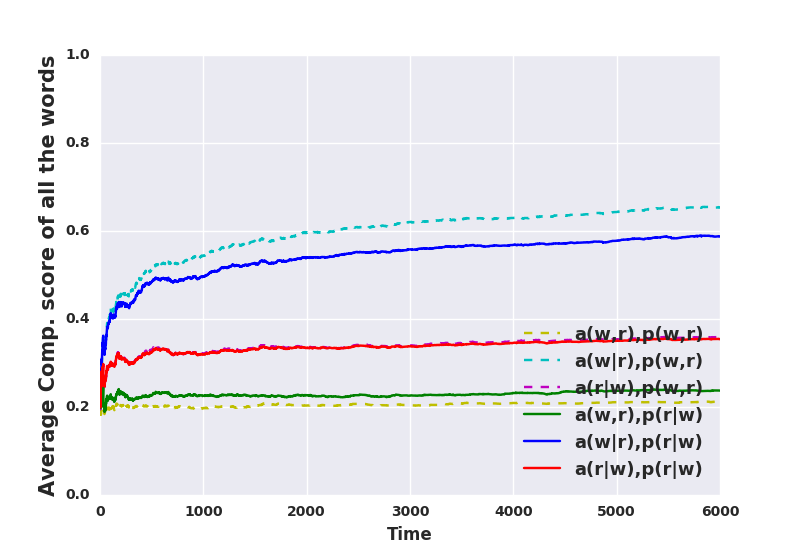}
\caption{LU+ input corpus.}
\label{fig:avgcomp-noru-lu2}
\end{subfigure}
\vspace{-0.2cm}
\label{fig:avgcomp}
\caption{(a) Average comprehension score of all models over 6K input pairs: (a) with no added referential/linguistic uncertainty; (b) with increased referential uncertainty (RU+) and (c) with linguistic uncertainty (LU+). In \Table{tb:models}, the implemented biases by each of these six models are reviewed.}
}
\end{figure}

We study the overall learning performance of each model by plotting the average comprehension score over time (\ie, 6,000 input pairs). The results are shown in \Figure{fig:avgcomp-noru-nolu}.
The best performing models (the dashed light blue and solid dark blue lines) are the models that implement word competition during in-the-moment learning (lines 2 and 5 of \Table{tb:models}). The latter model (the FAS model) also introduces referent competition into overall learning -- using $p(r|w)$ -- while the former does not have this competition -- using $p(w,r)$. Note that each competition type addresses a type of uncertainty: word competition reduces the linguistic uncertainty because the probability of \textit{irrelevant} words decreases as they compete with the relevant ones; similarly, referent competition limits the referential uncertainty. 
We expected that the best-performing model has mechanisms for addressing these two types of competition. However, our results show that the referent competition of calculating $p(r|w)$ during learning is not needed. 
The referent competition introduced during comprehension -- when a learned word meaning is compared to the gold-standard one -- is enough for the model using $a(w|r)p(w,r)$ to perform the same as $a(w|r)p(r|w)$.

The models that implement no competition in the alignment (lines 1 and 4 of \Table{tb:models})
have the worst performance (see the dashed light green and solid green lines); this shows the importance of competition during in the moment learning (local competition). We also observe that the two other models that only use referent competition in alignment (lines 3 and 6 of \Table{tb:models})
perform poorly. These models use the same type of competition during both learning and comprehension. This result suggests the importance of having both word and referent competitions occur at some point, during either learning or comprehension.

\subsubsection{The Role of Referential Uncertainty}

In \Figure{fig:avgcomp-ru2-nolu}, we see how the performance of each model changes as the level of referential uncertainty (RU) is increased. Recall in the RU+ corpus, there are referents in the scene representation that do not belong to any of the words from the utterance. We observe that the performance of all the models worsen when the referential uncertainty is increased. 
Interestingly, $a(w|r)p(w,r)$ performs better than $a(w|r)p(r|w)$ in the presence of increased referential uncertainty -- the decrease in their comprehension scores, compared to be trained over Input corpus, is 12\% and 21\%, respectively. This is interesting given that the latter implements referent competition in the word meaning representation and can limit referential uncertainty during learning. This result suggests that, to address RU, the referent competition during comprehension is best.

\subsubsection{The Role of Linguistic Uncertainty}

In \Figure{fig:avgcomp-noru-lu2}, we examine the effect of increased linguistic uncertainty (LU) in the overall performance of the models. Our results show that the performance of all models are decreased on the LU+ corpus. Here, we expect that the models with word competition to be more robust to LU. Indeed, we observe that $a(w|r)p(w,r)$, and $a(w|r)p(r|w)$ are the best models; the model that uses referent competition only during comprehension, $a(w|r)p(w,r)$ still has the best performance.
We also observe that the models that implement referent competition during in-the-moment learning (solid red and dashed purple lines) perform better than the ones without any competition (\ie, models using $a(w,r)$).
This suggests that some type of competition during in-the-moment learning is particularly important in addressing LU.

\subsubsection{The Role of Frequency}

\begin{figure}
\centering{
\begin{subfigure}{0.80\textwidth}
\includegraphics[width=\linewidth]{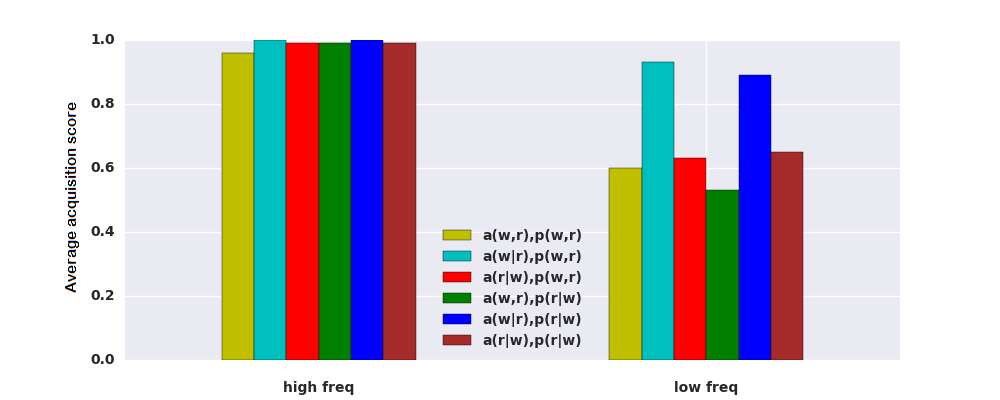}
\caption{Default input corpus: no added RU or LU.}
\label{fig:res-acq-freq}
\end{subfigure}\\
\begin{subfigure}{0.80\textwidth}
\includegraphics[width=\linewidth]{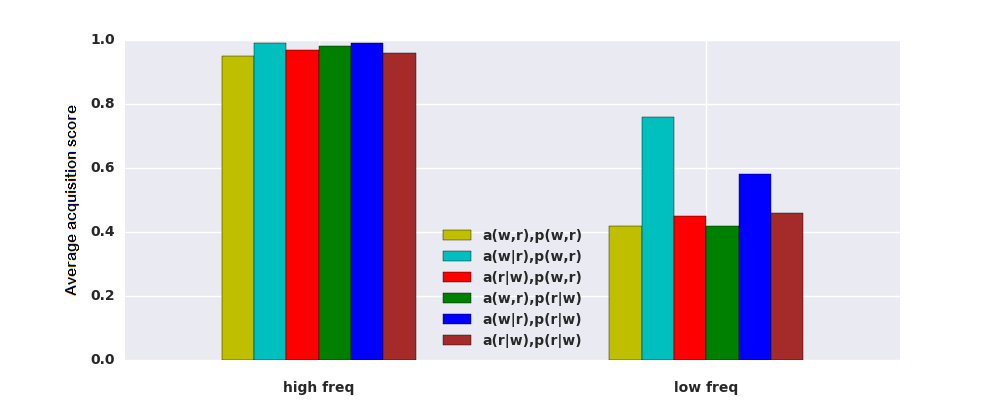}
\caption{RU+ input corpus.}
\label{fig:res-acq-freq-ru2}
\end{subfigure}\\
\begin{subfigure}{0.80\textwidth}
\includegraphics[width=\linewidth]{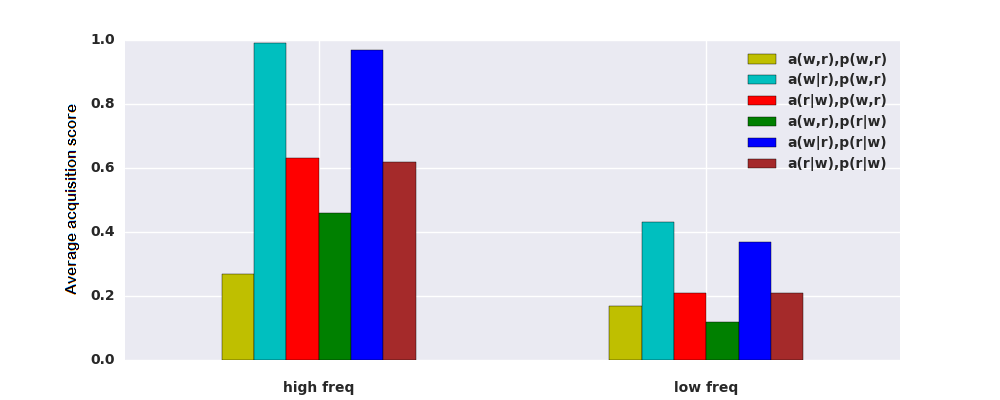}
\caption{LU+ input corpus.}
\label{fig:res-acq-freq-lu2}
\end{subfigure}
\vspace{-0.2cm}
\caption{Average comprehension score after 6K input items, split over word frequency.}
\vspace{-0.4cm}
}
\end{figure}

Low frequency words introduce another type of uncertainty because of the limited opportunity to gather evidence of their meaning, but children can learn the meaning of a word from only a few exposures \citep{markman.wachtel.1988,xu.tenenbaum.2007.psyrev}. Here we investigate how each model performs in such a situation. We categorize the words seen after $6$K inputs (with no added RU or LU) into two groups: \emph{low freq} ( 790 words seen less than 5 times) and \emph{high freq} (240 words seen more than 10 times). See \Figure{fig:res-acq-freq}. 

For high-freq words, all the models perform very well -- their average comprehension score is equal to or higher than $0.96$. For low-freq words, we observe a similar pattern as before; again, the best-performing model is $a(w|r)p(r,w)$ (average comprehension score of $0.93$) and the second best model is the FAS model  ($a(w|r)p(r|w)$, with the score of $0.89$).  The other four models perform very similarly, but there is a large gap between their average comprehension score and the two best models (at least $0.24$).
These results are particularly interesting because they are in line with empirical observations that suggest the mutual exclusivity bias is important in guiding children in learning from a few exposures. In our simulations, the models that implement this bias (word competition in the moment of learning) can learn word meanings from only a few examples.

Next we explore how the interaction of word frequency and referential/linguistic uncertainty affect the comprehension score; see \Figure{fig:res-acq-freq-ru2} and \Figure{fig:res-acq-freq-lu2}.
As expected, we observe that the average comprehension scores of all models decrease as the input contains more uncertainty -- i.e., with added RU or LU. We also observe that the models are more robust when RU increases; for most models, the average decrease in comprehension score is greater in the presence of LU (as opposed to RU). 
Interestingly, under increased referential uncertainty, all models learn the high-freq words, but under increased linguistic uncertainty, only those with in-the-moment word competition ($a(w|r)p(w,r)$ and $a(w|r)p(r|w)$) achieve high comprehension scores. 

This discrepancy in performance given the type of uncertainty is a result of the nature of the word learning problem itself: there are two sources of competition (between referents or words) and three stages in which the competition can happen (during in-the-moment, overall learning, and comprehension). However, the word competition  -- that addresses the linguistic uncertainty -- only happens \emph{locally} for each observation during the in-the-moment learning.
First, because we model learning a meaning representation for each word (and not a ``word representation'' for each referent), the learned representations only encode competition (if any) over referents for a given word. Thus, referents (unlike words) can compete globally -- all the observed referents (as opposed to the referents in the current observation) compete for a given word.
This global competition imposes a stronger mutual exclusivity bias that better addresses referential uncertainty: it is easier for the model to ignore irrelevant referents because it has a tendency to associate a given word to only one referent, the one that has consistently and frequently co-occurred with that word.
Also, as noted earlier, the competition during comprehension is again only between referents, because we use cosine to compare the learned and gold-standard representation of words.

\subsubsection{Learning Synonyms and Homonyms}
\label{sec:LearningSynonymsAndHomonyms}

Although evidence suggests that children have a mutual exclusivity bias that disprefers synonyms and homonyms \citep{markman.wachtel.1988,casenhiser2005children}, given sufficient exposure, children eventually overcome this bias to learn multiple words associated with an identical (or similar) meaning, or to learn single words associated with multiple distinct meanings.
When studying children's ability in learning homonymous words, to control a child's first exposure to the second meaning of a known word, a pseudo-homonym is used. Analogously, pseudo-synonyms are used to study acquisition of synonymous words. A pseudo-homonym is a familiar word to the learner (\eg, \textit{ball}) but it is used to convey a novel meaning (\eg, \textsc{dax}). Similarly, a pseudo-synonym is used in studying the acquisition of a second label (e.g., \textit{dax}) for a familiar referent (e.g., \textsc{ball}). 
Here, to study the performance of the models in learning homonymy and synonymy, we use pseudo-homonyms and pseudo-synonyms as well.  We only report the results of the two best performing models, $a(w|r)p(w,r)$ and $a(w|r)p(r|w)$.
We observe that all models are successful in learning synonyms -- can associate two words with one referent; however, only models that represent overall meanings as a joint distribution (\ie, $p(w,r)$) can associate two referents with one word. The competition among referents, introduced by $p(r|w)$, for a given word, prevents two referents to be simultaneously associated with one word.

\begin{figure*}[!th]
\begin{subfigure}{.48\textwidth}
\includegraphics[width=\linewidth]{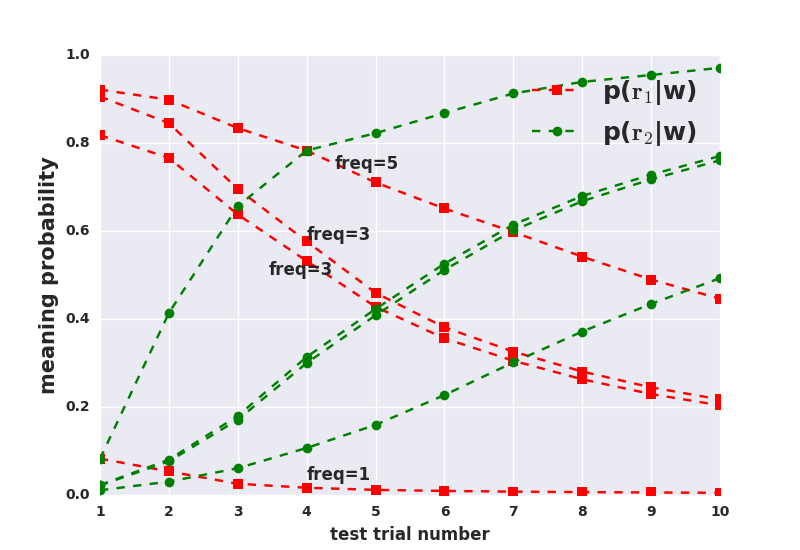}
\caption{Low frequency}
\label{fig:faslowfreq}
\end{subfigure}
\begin{subfigure}{.48\textwidth}
\includegraphics[width=\linewidth]{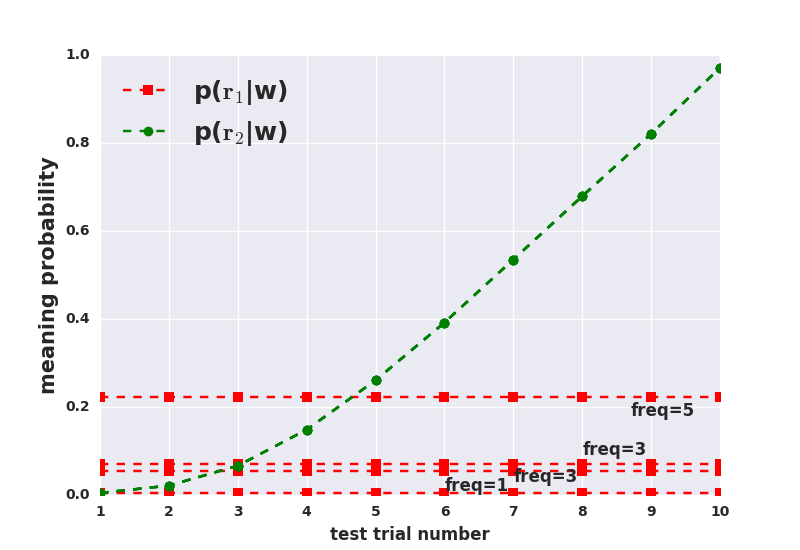}
\caption{Low frequency}
\label{fig:wclowfreq}
\end{subfigure}\\
\begin{subfigure}{.48\textwidth}
\includegraphics[width=\linewidth]{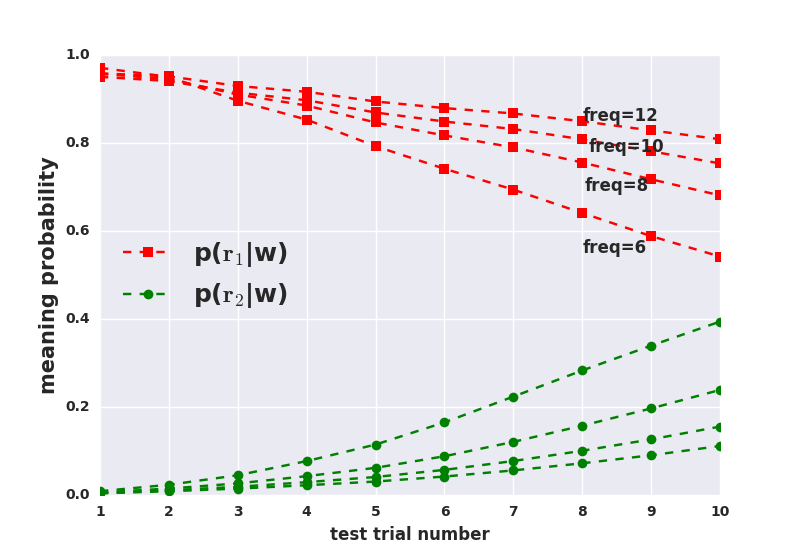}
\caption{Medium frequency}
\label{fig:fasmidfreq}
\end{subfigure}
\begin{subfigure}{.48\textwidth}
\includegraphics[width=\linewidth]{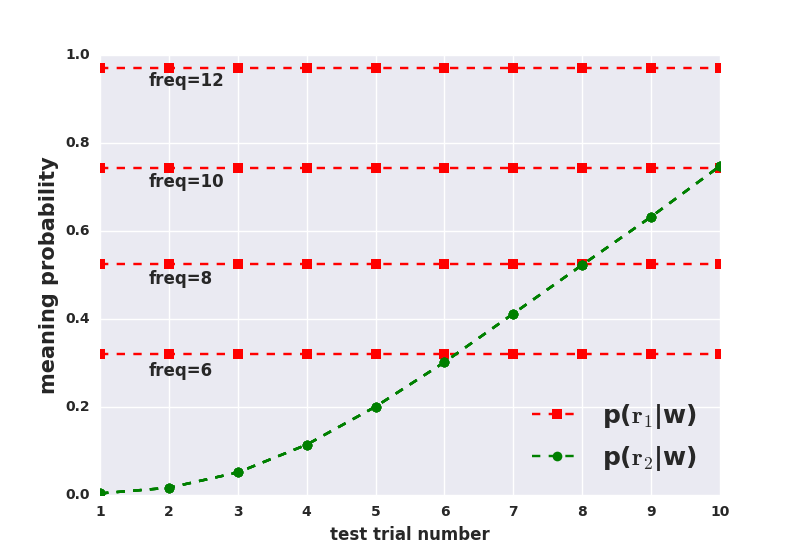}
\caption{Medium frequency}
\label{fig:wcmidfreq}
\end{subfigure}\\
\begin{subfigure}{.48\textwidth}
\includegraphics[width = \linewidth]{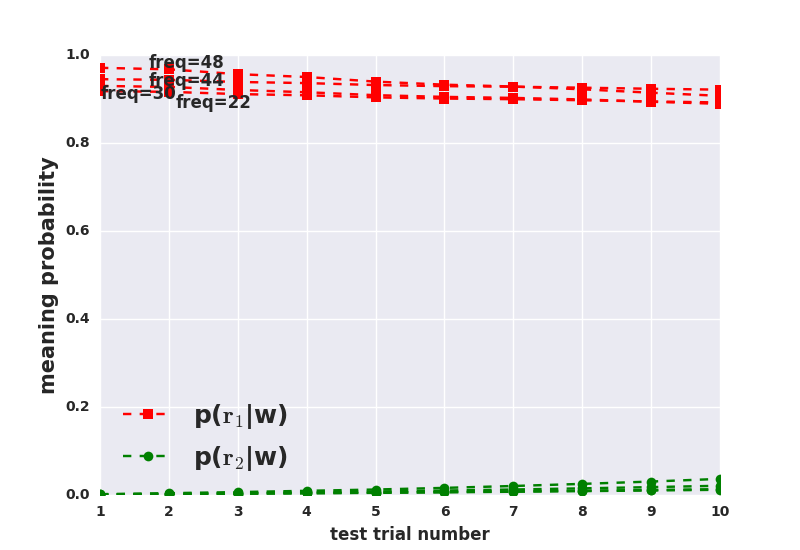}
\caption{High frequency}
\label{fig:fashighfreq}
\end{subfigure}
\begin{subfigure}{.48\textwidth}
\includegraphics[width = \linewidth]{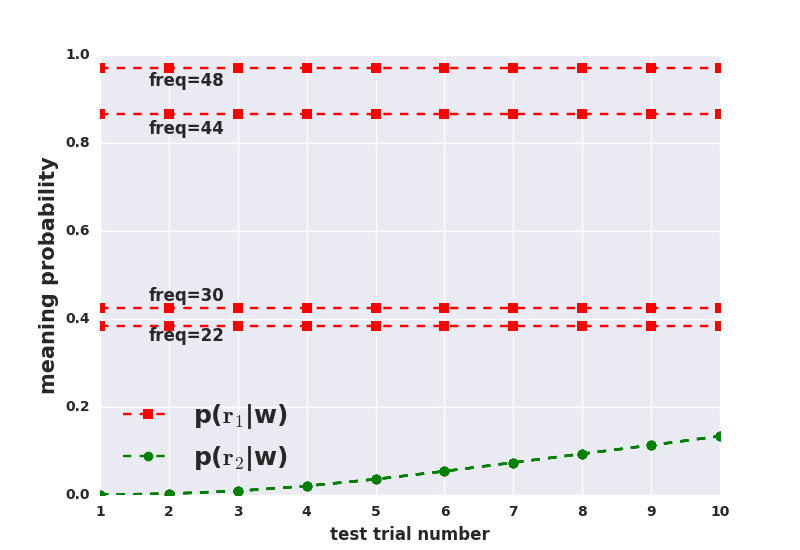}
\caption{High frequency}
\label{fig:wchighfreq}
\end{subfigure}\\
\vspace{-0.2cm}
\caption{Patterns of change in the meaning probabilities of the first and second meaning of pseudo-homonyms in the FAS model ($a(w|r)p(r|w)$, left column) and $a(w|r)p(r,w)$ (right column). The \textbf{freq} is the number of times the first meaning is observed by the model.}
\label{fig:fas_homonyms}
\end{figure*}

\paragraph{Learning Homonymous Words.} Similarly to \citet{fazly.etal.2010.csj}, we train all models on 1000 input pairs, and then present the models with 10 test trials. In each of these 10 trials, we include a pseudo-homonym -- a familiar word added to an utterance, paired with a novel meaning added to the utterance's corresponding scene representation. 
These 10 utterance-scene pairs are chosen such that they provide a context that is relevant to the new meaning but irrelevant to the word's first meaning. We choose 4 familiar words from each of 3 frequency ranges (less than 5, between 5 and 20, and more than 20), to study words whose association to their first meaning is more or less strong.\footnote{By test trial, we mean after 1000 input pairs, we added 10 other input pairs. These 10 input pairs keep a same context for the new meaning. The 3x4 pseudo-homonyms refer to the point that we run this setup and experiment 12 times for 12 words which are included in 3 categories of less than 5, between 5 and 20, and more than 20.}

The $a(w|r)p(r,w)$ model can learn the second meaning of the words (green line); the comprehension score of the second meaning, increases as the model receives more examples of the second meaning (see \Figure{fig:wclowfreq}, \Figure{fig:wcmidfreq}, and \Figure{fig:wchighfreq}).  This happens because there is no competition enforced among the referents for a word during overall learning. 
Moreover, the comprehension score of the first meaning is not affected as the model learns the second meaning.
We observe the opposite pattern with the FAS model: as the model observes the second meaning of the word, the probability of the first meaning decreases, due to the competition in \Equation{eq:probfgivenw}. (See \Figure{fig:faslowfreq}, \Figure{fig:fasmidfreq}, and \Figure{fig:fashighfreq}; cf. \citep{fazly.etal.2010.csj}.)

\begin{figure*}[!th]
\begin{subfigure}{.48\textwidth}
\includegraphics[width=\linewidth]{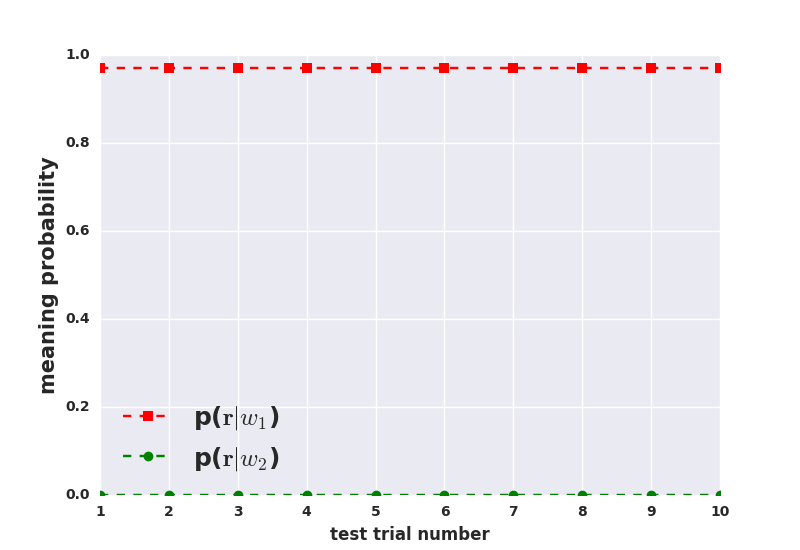}
\caption{$a(w,r)p(w,r)$}
\label{fig:asyn}
\end{subfigure}
\begin{subfigure}{.48\textwidth}
\includegraphics[width=\linewidth]{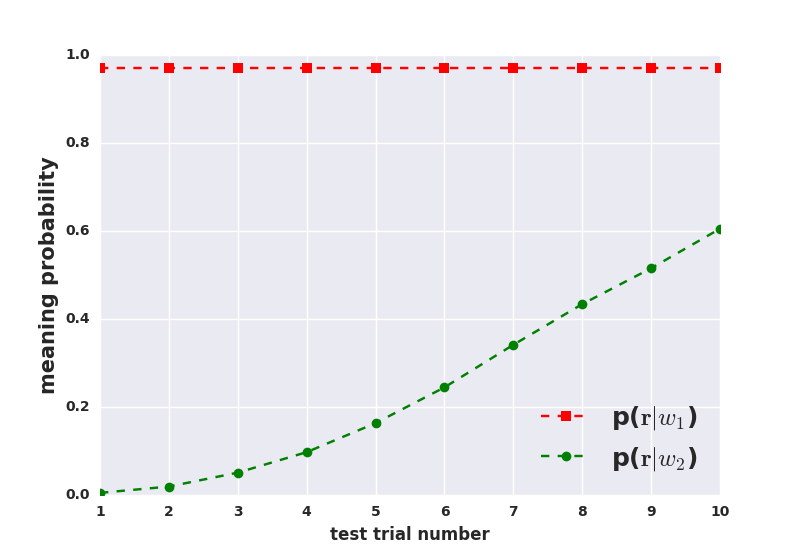}
\caption{$a(w|r)p(w,r)$}
\label{fig:bsyn}
\end{subfigure}\\
\begin{subfigure}{.48\textwidth}
\includegraphics[width=\linewidth]{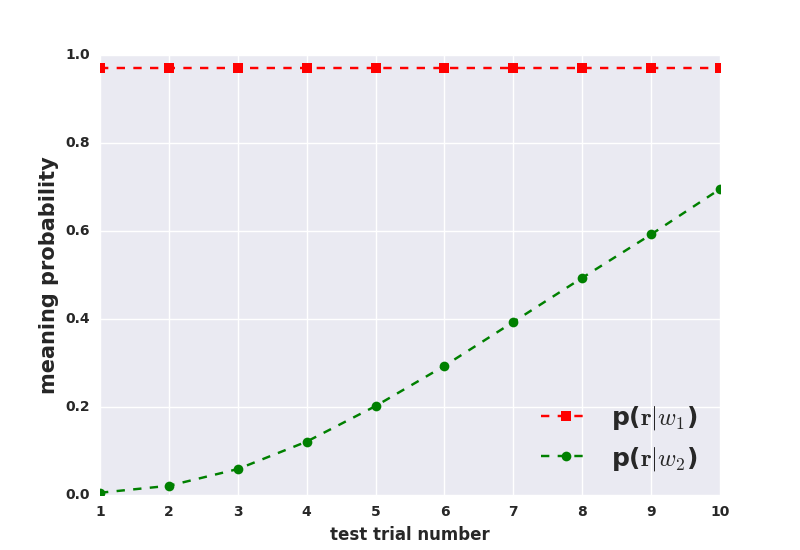}
\caption{$a(r|w)p(w,r)$}
\label{fig:csyn}
\end{subfigure}
\begin{subfigure}{.48\textwidth}
\includegraphics[width=\linewidth]{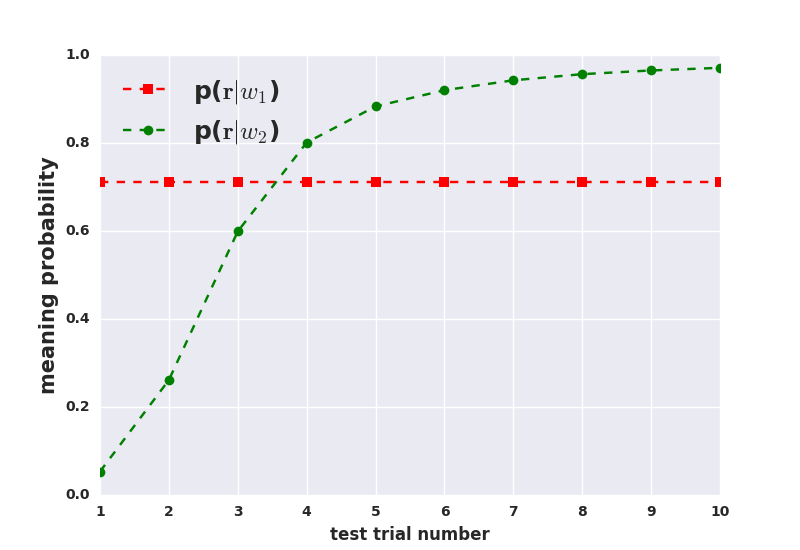}
\caption{$a(w,r)p(r|w)$}
\label{fig:gsyn}
\end{subfigure}\\
\begin{subfigure}{.48\textwidth}
\includegraphics[width=\linewidth]{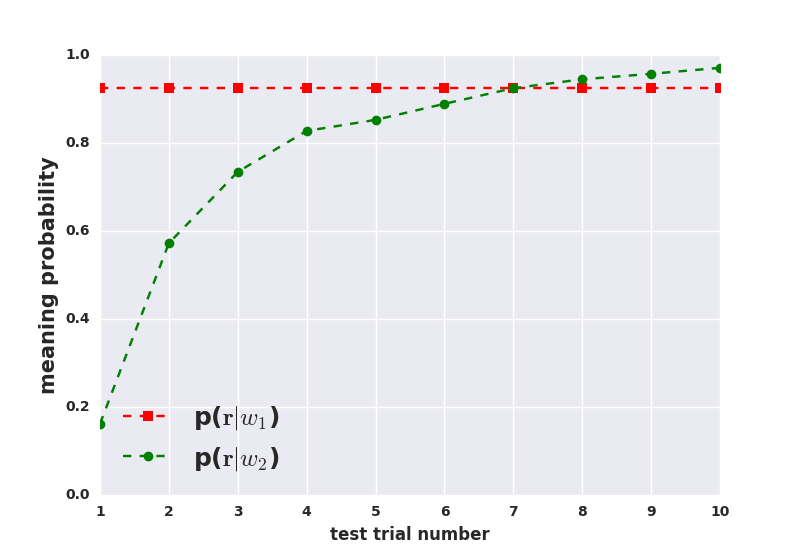}
\caption{$a(w|r)p(r|w)$}
\label{fig:fassyn}
\end{subfigure}
\begin{subfigure}{.48\textwidth}
\includegraphics[width = \linewidth]{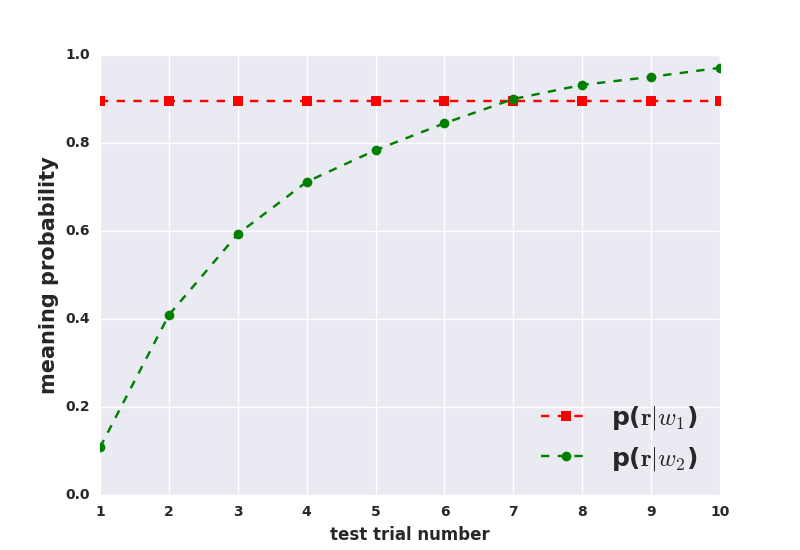}
\caption{$a(r|w)p(r|w)$}
\label{fig:isyn}
\end{subfigure}
\vspace{-0.2cm}
\caption{Patterns of change in the meaning probabilities of the first and second word of pseudo-synonyms in all model.}
\label{fig:all_synonyms}
\end{figure*}

\paragraph{Learning Synonymous Words.} 
As mentioned earlier, given sufﬁcient training, children can acquire the meaning of a new label for a meaning already associated with a known word. 
We replicate the experiments of \citet{fazly.etal.2010.csj} on learning synonyms.
We run 20 simulations for each modeling condition, where a simulation consists of training the model followed by a set of 10 test trials. The results presented in \Figure{fig:all_synonyms} show the average meaning probabilities over the 20 simulations.
We observe that all the models (except $a(w,r)p(w,r)$\footnote{For model $a(w,r)p(w,r)$, it earns values for the second label score as well, but because the scores for first label are not normalized (in \Equation{eq:probwandf}), they are large values. So, getting a score for the second label seems like 0 compared to that, not actual zero.}) successfully learn to associate a second word to a familiar referent.
Moreover, the learning of the second word does not affect the comprehension score of the first word associated with that label. These results are expected as none of the models enforce a word competition during the overall learning of the word meaning representation. See \Figure{fig:all_synonyms}.

\vspace{-.1cm}

%% file: 050-conclusions.tex
\section{Conclusion}
The computational level of analysis \citep{marr.1982} allows us to contemplate what problem a cognitive phenomenon solves.
For example, learning word meanings via cross-situational statistics can be formulated as finding mappings between words and referents that are most consistent with our observations.
On the other hand, modeling cognition at the algorithmic level \citep{marr.1982} plays an important role in providing insight about cognitive mechanisms; it requires specifying the details of algorithms and representations which in turn enables us to study their role and interaction at different stages of learning.

Previous research has studied word learning at both algorithmic and computational levels \citep[\eg,][]{siskind.1996,yu.ballard.2007,frank.etal.2007,fazly.etal.2010.csj}. We proposed a framework for modeling cross-situational word learning at the computational level that unifies some of previous work in this domain -- approaches that formulate word learning as a translation problem.
We also show that instantiating this framework results in different word learning models at the algorithmic level: each model has specific inductive biases that define how words and referents compete for association strength given a word/referent. Our framework enables a thorough comparison of these models with respect to a range of word learning behavior such as modeling of homonyms and synonyms.

More specifically, at the algorithmic-level of analysis, we examine how competition among words or referents plays a role in:
(1) learning from a given observation or in-the-moment learning, (2) overall learning of word meanings, and (3) comprehension of a given word. Moreover, we investigate how these assumptions change the performance of a model in the face of uncertainty.
Our results show that models that implement the two complementary types of referent and word competition perform the best. Each competition type addresses a specific type of uncertainty -- word and referent competitions address linguistic and referential certainty, respectively; because the word learning input has both uncertainties, it is important for a model to implement the two competitions. 
These models are the most robust and learn successfully from few examples. 
Moreover, we find that the best model implements competition during in-the-moment learning and comprehension, but not a global competition over word meaning representations.  By avoiding an overall word meaning competition, the model is able to successfully learn multiple meanings of ambiguous words, given sufficient evidence.

Through considering different algorithmic-level implementations of the same computational-level theory, we show that the well-known mutual exclusivity bias can address the uncertainty of the environment at different stages of learning. 
Moreover, we discovered a model formulation that better predicts observed word learning data, particularly it can learn homonyms without overriding the competition processes that implement the mutual exclusivity bias. Our work highlights the need for validating a computational-level theory through a thorough algorithmic-level analysis. Such analysis will make the assumptions behind a model explicit, which is needed to fully understand the processes underlying a cognitive phenomena.

%% file: 060-appendix.tex
\appendix 

\section{The Derivation of the FAS model}
\label{sec:appendix}
The FAS model assumes that referents are generated independently given an utterance $U$; instead of calculating $p(s,u|\theta)$ as in \Equation{eq:cll}, the likelihood is defined over the conditional probability of referents given an utterance:
\begin{align}
\vspace{-.2cm}
p(r|u, \theta) &= \sum_{a \in A} p(r, a|u, \theta) 
\vspace{-.2cm}
\end{align}
Here, the alignment variable defines the mappings between words to a given referent. More specifically, the value of the alignment variable $a$ selects the word in the utterance $u$ (\eg, $w_j$) that is mapped to a given referent $r_i$.
\begin{align}
\vspace{-.2cm}
p(a_{j}|u,r_i,\theta) &= \frac{p(a_{j},r_i| u, \theta)}{p(r_i|u, \theta)} = \frac{\theta_{ij}} {\sum_{w_k \in u} \theta_{ik}} 
\vspace{-.2cm}
\end{align}
where $\theta_{ij}$ returns the association of $w_j$ to the referent $r_i$ given the learned distribution $\theta$. Note that this corresponds to the Expectation step of the EM algorithm, and \Equation{eq:emfasalign} is an instantiation of \Equation{eq:emalign}. 
\break
\break
In the Maximization step, the new value of $\theta$ is calculated by finding $\theta$ that maximizes the model likelihood:
\begin{align}
\textrm{E}_{A|C, \theta^{t-1}}[\log p(S, A|U, \theta)]
\end{align}
where $S$ and $U$ are a set of all scenes and utterances, respectively:
\begin{align}
\vspace{-.2cm}
\textrm{E}_{A|C, \theta^{t-1}}[\log p(S, A|U, \theta)]
&=\sum_{a \in A} \log\big(\prod_{\substack{u \in U \\ s \in S}} p(s,a |u, \theta) \big)p(a|C, \theta^{t-1})
\nonumber\\ 
&=\sum_{a \in A} \log\big(\prod_{\substack{u \in U \\ s \in S}} \prod_{\substack{r \in s}} p(r,a |u, \theta) \big)p(a|C, \theta^{t-1})
\nonumber\\ 
&=\sum_{a \in A} \sum_{\substack{u \in U \\ r \in S}} \log p(r,a |u, \theta)p(a|C, \theta^{t-1})
\nonumber \\
&=\sum_{\substack{w_j \in U \\ r_i \in S}} \log  p(r_i |w_j, \theta) \mathrm{count}(r_i, w_j)p(a_j|u, r_i, \theta^{t-1})
\nonumber \\
&=\sum_{\substack{w_j \in U \\ r_i \in S}} \log \theta_{ij} \mathrm{count}(r_i, w_j)\ p(a_j|u, r_i, \theta^{t-1})
\label{eq:mstepfas}
\vspace{-.2cm}
\end{align}
where $j$ (determined by the alignment variable $a$) is the word mapped to $r_i$, and $count(r_i,w_j)$ is the number of times $r_i$ and $w_j$ have co-occurred in the corpus $C$.

The FAS model assumes that $\theta$ is a conditional probability, $p(r|w)$. This means that there is an additional dependence assumption on the learned representations: each word is a distribution over features, and thus given a word, the features compete to be associated with that word.
To impose this new assumption on the representation, a constraint is added to the expectation defined in \Equation{eq:mstepfas}:
\begin{align}
\vspace{-.2cm}
\Lambda(\theta, \lambda) = 
\sum_{\substack{w_j \in U \\ r_i \in S}} \log \theta_{ij} \ \mathrm{count}(r_i, w_j) \ p(a_j|u, r_i, \theta^{t-1})\\
- \sum_{w_m \in U}\lambda_{w_m}(\sum_{r_n \in S}p(r_n|w_m) - 1) 
\vspace{-.2cm}
\end{align}
Note that the Lagrange multipliers $\lambda_{w_m}$ ensures that the new constraint on $\theta$ -- a distribution over referents for each word -- is satisfied. 

To find $\theta$ that maximizes the expectation in \Equation{eq:mstepfas}, the derivative of objective function $\Lambda(\theta, \lambda)$ is calculated and equated to zero: 
\begin{align}
\vspace{-.2cm}
\frac{\partial\Lambda}{\partial p(r_i|w_j)} = 
\frac{p(a_j|u,r_i, \theta^{t-1})\ \mathrm{count}(r_i, w_j)}{\theta_{ij}} - \lambda_{w_m} = 0 \nonumber\\
\Rightarrow p(r_i|w_j)= 
p(a_j|u,r_i, \theta^{t-1})\ \mathrm{count}(r_i, w_j) \frac{1}{\lambda_{w_m}} 
\vspace{-.2cm}
\end{align}
Given $\sum_{r_n}p(r_n|w_m)= 1$, we calculate $\lambda_{w_m}$:
\begin{align}
\vspace{-.2cm}
 \Rightarrow \lambda_{w_m} = \sum_{r_n \in S} 
 p(a_j|u,r_i, \theta^{t-1}) \ \mathrm{count}(r_i, w_j)
\vspace{-.2cm}
\end{align}
Using the above $\lambda_w$ and \Equation{eq:emfasalign} to calculate the alignment probabilities, we have:
\begin{align}
\vspace{-.2cm}
p(r_i|w_j)= \frac { p(a_{j}|u,r, \theta^{t-1})\ \mathrm{count}(r_i, w_j)} 
{\sum\limits_{r_m \in S}  p({a_j}|u,r_m, \theta^{t-1})\ \mathrm{count}(r_m, w_j)}
\vspace{-.2cm}
\end{align}
We can approximate $p(a_{j}|u,r, \theta^{t-1})\ \mathrm{count}(r_i, w_j)$ by adding the current alignment probability, $p(a_{j}|u,r, \theta^{t-1})$, to the sum of all the previously calculated ones (instead of multiplying it to the number of times the word and referent co-occur). This approach is an approximation of $p(a_{j}|u,r, \theta^{t-1})\ \mathrm{count}(r_i, w_j)$ because the value of the alignment probability changes after processing each $u$-$s$ pair, but it can be calculated incrementally; FAS defined an association score, $assoc$ which is updated as the model process $u$-$s$ pairs,
\vspace{-.1cm}
\begin{eqnarray}
\mathrm{assoc}_{t}(w_j,\,r_i) = \mathrm{assoc}_{t-1}(w_j,\,r_i) +  a_t(w_j|r_i)
\vspace{-.2cm}
\end{eqnarray}
where $a_t(w_j|r_i)=p({a_{j}}|u,r_i, \theta^{t-1})$ and the initial value of $\mathrm{assoc}(w_j,\,r_i)$ (before the first co-occurrence of $w_j$ and $r_i$) is zero.  
Intuitively, this score shows the overall association strength of a word and a referent and it captures how strongly the word-referent pair are associated in each observation, $u$-$s$.